\documentclass[10pt,twocolumn,letterpaper]{article}

\usepackage{cvpr}              

\usepackage{graphicx}
\usepackage{amsmath}
\usepackage{amsthm}
\usepackage{amssymb}
\usepackage{booktabs}
\usepackage{multirow}

\usepackage[pagebackref,breaklinks,colorlinks]{hyperref}

\usepackage[capitalize]{cleveref}

\DeclareMathOperator*{\argmin}{argmin}

\crefname{section}{Sec.}{Secs.}
\Crefname{section}{Section}{Sections}
\Crefname{table}{Table}{Tables}
\crefname{table}{Tab.}{Tabs.}

\def\cvprPaperID{*****} 
\def\confName{CVPR}
\def\confYear{2022}

\begin{document}

\title{Unseen Object 6D Pose Estimation: A Benchmark and Baselines}

\author{Minghao Gou$^{1,2}$\thanks{Equal contribution}~\thanks{Work done as an intern in Alibaba XR Lab}, Haolin Pan$^{1*}$, Hao-Shu Fang$^{1}$, Ziyuan Liu$^{2}$, Cewu Lu$^{1}$\thanks{Cewu Lu is the corresponding author}, Ping Tan$^{2,3}$\\
 $^{1}$Shanghai Jiao Tong University, $^{2}$Alibaba XR Lab, $^{3}$Simon Fraser University\\
{\tt\small gmh2015@sjtu.edu.cn,  01212015131405@sjtu.edu.cn, fhaoshu@gmail.com,}\\
{\tt\small ziyuan-liu@outlook.com, lucewu@sjtu.edu.cn, pingtan@sfu.ca}\\
}

\maketitle

\begin{abstract}
Estimating the 6D pose for unseen objects is in great demand for many real-world applications. However, current state-of-the-art pose estimation methods can only handle objects that are previously trained. 
In this paper, we propose a new task that enables and facilitates algorithms to estimate the 6D pose estimation of novel objects during testing.
We collect a dataset with both real and synthetic images and up to 48 unseen objects in the test set.
In the mean while, we propose a new metric named Infimum ADD (IADD) which is an invariant measurement for objects with different types of pose ambiguity.
A two-stage baseline solution for this task is also provided. By training an end-to-end 3D correspondences network, our method finds corresponding points between an unseen object and a partial view RGBD image accurately and efficiently. It then calculates the 6D pose from the correspondences using an algorithm robust to object symmetry.
Extensive experiments show that our method outperforms several intuitive baselines and thus verify its effectiveness.
All the data, code and models will be made publicly available. Project page: \url{www.graspnet.net/unseen6d}

\end{abstract}

\section{Introduction}\label{sec:introduction}
\def\ECCVSubNumber{1671}
\newcommand{\pose}{6D pose}
\newcommand{\Pose}{6D Pose}
\newcommand{\segmethodfull}{Object-Level Segmentation Proposal Network}
\newcommand{\correpmethodfull}{Object-Scene Correspondence Network}
\newcommand{\segmethod}{SPN}
\newcommand{\correpmethod}{OSCN}
\newcommand{\task}{unseen object 6D pose estimation}
\newcommand{\method}{3DCor}
\newcommand{\metric}{IADD}
\newcommand\minghao[1]{\textcolor{red}{Minghao:#1}}
\newcommand*\phl[1]{\textcolor{blue}{{Haolin:}#1}}

Object \pose\ estimation is an important task in computer vision and robotics. Many real-world applications \cite{OCRTOC,amazonpick} such as grasping and VR/AR heavily rely on accurate object \pose\ estimation result.

Researches on object pose estimation have been explored for a long time.
Template-based methods such as point cloud registration~\cite{icp,yang2016fast} and template matching~\cite{jurie2002real,drost2010model} mainly adopt handcrafted rules to encode geometry features. Since the geometry encoding schema is model agnostic, these methods are applicable to any objects in principle. However, due to the inferior expressive power of handcrafted features, they cannot achieve satisfactory results in cluttered scenes with noise and need many manual tuning efforts. 
Recently, deep learning methods based on 2D image~\cite{ssd6d,bb8,posecnn,pvnet} or 3D point cloud~\cite{densefusion,pvn3d,ffb6d} are proposed to tackle this problem and yield better performances, benefiting from the powerful feature extraction ability of neural network.


However, in the current task setting of 6D pose estimation~\cite{bop,linemod,posecnn}, the same object set is shared in both training and testing phase. Taking such assumption that the testing object is always available during the training period, current state-of-the-art 6D pose estimation algorithms~\cite{ffb6d,pvn3d} follow the schema that directly models the object's texture and geometry features within the neural networks. Prior knowledge of object models such as keypoint location~\cite{pvnet,pvn3d} or voting offsets~\cite{pvnet,pvn3d} is also encoded by the networks. It turns out that these methods can only estimate the \pose\ of known objects during training. In real-world applications such as the flexible robotic assembly, novel objects appear frequently. To detect their \pose s, new data collection process including keypoints allocation and synthetic image generation~\cite{2017Cut} needs to be repeated, and the network needs to be retrained. This is labor intensive and prevents the 6D pose estimation algorithms from rapid deployment.

\begin{figure*}[tbp]
    \centering
    \includegraphics[width=0.9\linewidth]{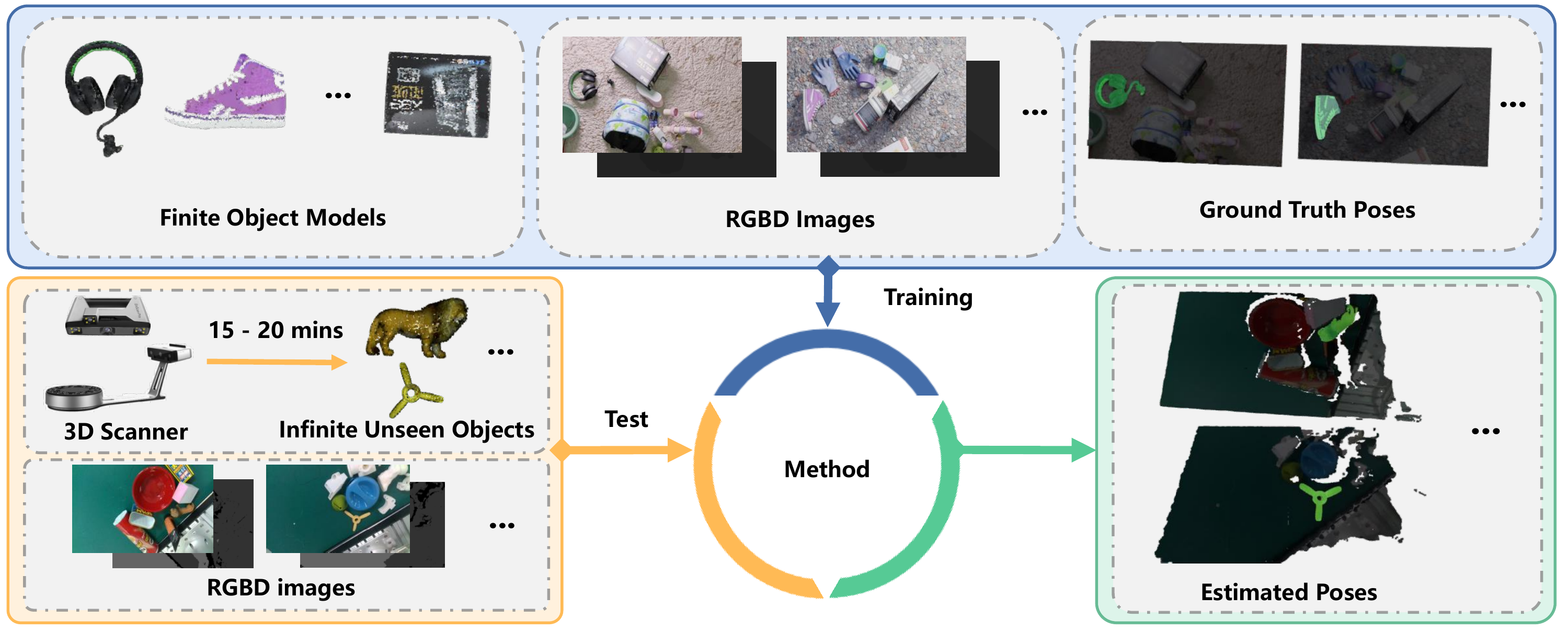}
    \caption{Illustration of the task of Unseen Object 6D Pose Estimation}
    \label{fig:introduction}
\end{figure*}

In this paper, we reconsider this problem and propose to explore a new direction. In practice, the mesh model of an object is easy to obtain. With a commercial 3D scanner, the mesh model of an object can be retrieved within minutes. The major bottlenecks for fast deployment of the aforementioned methods are the synthetic data generation and network retraining processes.
Thus, as shown in Fig.~\ref{fig:introduction}, we propose a new task named \textit{\task}. After training on a finite set of objects, the algorithm is required to estimate the \pose\ of any novel object in a scene given their mesh models but without re-training.
This task is similar to the original \pose\ estimation problem except that the mesh models of objects in the test set will not be available during training.

To fulfill the task, we propose a new benchmark that contains a training set with over 1000 objects and 1500 scenes and a test set with 48 novel objects and 90 real-world captured scenes, built on top of~\cite{graspnet,google1000,blenderproc}. We also propose a two-stage baseline solution. In the first stage, a 3D correspondences detection network takes an object mesh model and a single-view partial scene point cloud as inputs. Its goal is to segment the object from the scene and detect dense 3D correspondences sequentially. This network is trained in an end-to-end manner with multi-task losses. In the second stage, we follow EPOS~\cite{epos} to calculate \pose\ given the dense correspondences.


To better evaluate the performances of different algorithms in the future, we propose a new metric named \metric\ for our benchmark. It overcomes the limitations of previous metrics and is capable of evaluating the \pose\ estimation accuracy of any object in a unified manner even if the object has infinite pose ambiguities.


We conduct extensive experiments on our benchmark to verify the effectiveness of our method. Several intuitive baselines for this task are carefully implemented for comparison. Our algorithm achieves \textbf{20.3}, \textbf{14.9} and \textbf{9.7} Area Under Curve (AUC) improvements over carefully tuned baselines on three test subset with different kinds of objects respectively. We also evaluate all methods on the YCB-Video~\cite{posecnn} test set without retraining, and a \textbf{11.2} AUC improvements over our implemented baselines is witnessed.




The main contributions of this paper are as follows:
\vspace{-0.1in}
\begin{itemize}
    \item We propose a new task of which the \pose\ estimation algorithms can transfer to unseen objects easily.
    \item We present a new benchmark with sufficient training data, real-world test set and a new metric on \pose\ estimation. With the standard dataset and evaluation metric, we hope to facilitate the fair comparison of different future methods.
    \item We develop a baseline solution for this task, which implements a framework of instance level segmentation and \pose\ estimation for unseen objects. 

\end{itemize}
\section{Related Works}\label{sec:relatedworks}
In this section, we briefly review previous researches on object \pose\ estimation, 3D correspondence, and \pose\ estimation metrics.

\subsection{Object \Pose\ Estimation Algorithms} \label{subsec:relatedworks_algorithms}
Current existing algorithms can be divided into mainly three types.

\begin{figure}[t]
    \centering
    \includegraphics[width=0.95\linewidth]{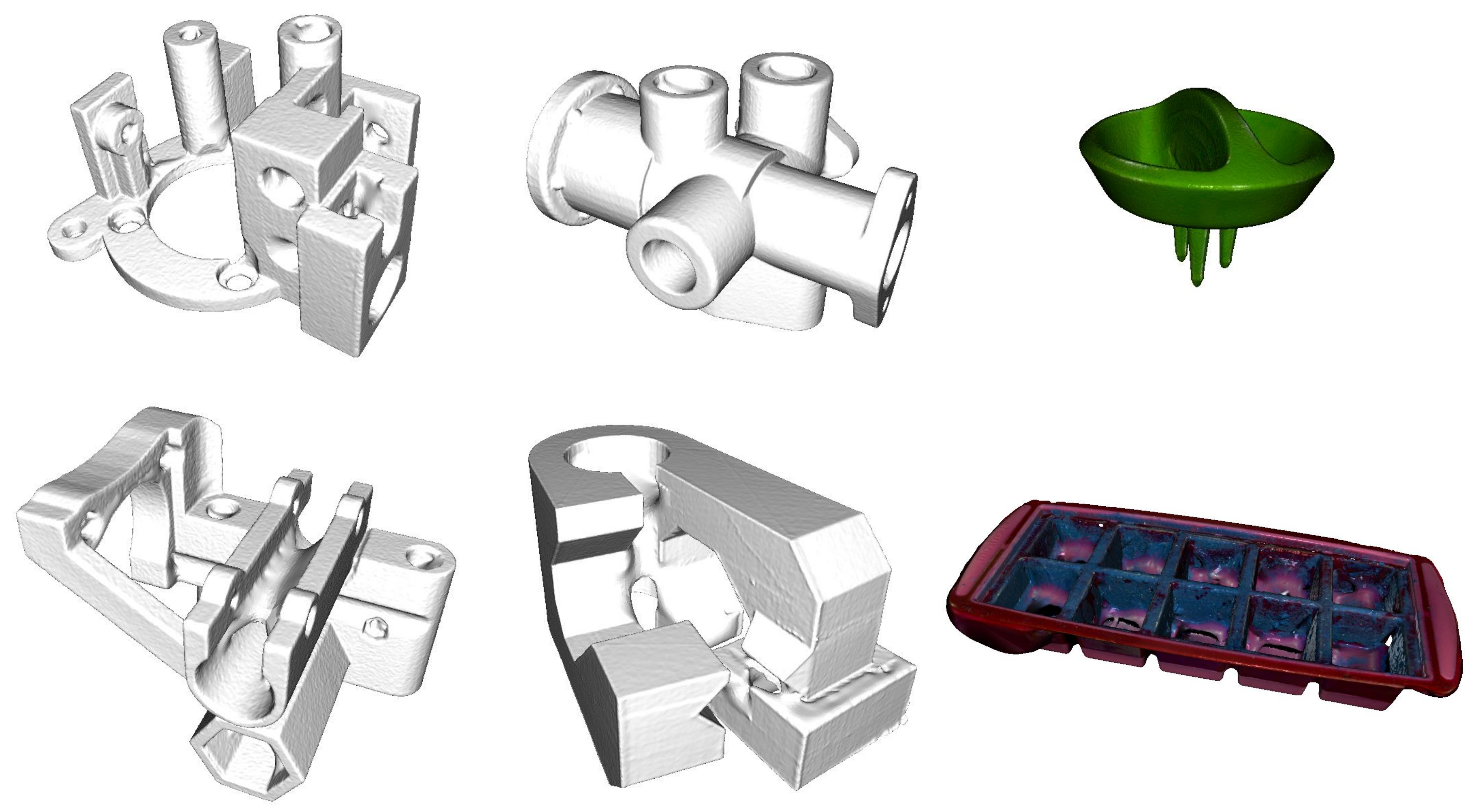}
    \caption{Examples of the unseen objects in the test set.}
    \label{fig:vis}
\end{figure}

\noindent\textbf{Pose prediction methods} tend to directly obtain the object \pose\ from image features. Some~\cite{ssd6d,posecnn,densefusion,trabelsi2021pose} apply classification or regression to get the object \pose\ after extracting pattern features by deep neural networks. Others~\cite{latentfusion,deepim} iteratively optimize the object \pose\ by minimizing the re-projection error. These algorithms work well when objects are with rich texture but fail on texture-less objects or occlusions.  Other methods \cite{maae} requires an additional 2D detector. \cite{nocs} focuses on category level pose estimation and is not suitable for our task, since the test objects are totally novel (see Figure~\ref{fig:vis}). 

\noindent\textbf{Correspondences based methods} aim to firstly detect 2D or 3D object keypoints in the image and then solve a PnP or fitting problem to obtain the object \pose.\ In the former case, methods extract 2D keypoints~\cite{bb8,zhao2018estimating,okpose,pvnet} of the target objects and apply PnP-RANSAC~\cite{ransac} algorithms to obtain their 6D poses. The latter ones~\cite{pvn3d,ffb6d} find 3D keypoints of the target objects and calculate the \pose\ by least square fitting. These methods require a definition of the keypoints as prior knowledge.

Beyond the former types of methods that require the network to implicitly remember the target objects during training,
\noindent\textbf{registration based methods}~\cite{icp,lee2021deep,wang2019deep,prnet,yang2020mapping, 3dmatch,scan2cad} treat this task as point cloud registration and could estimate the 6D pose between two novel inputs. However, they usually consider the registration between two similar-sized targets. In our cases, the intersection of union(IoU) between the mesh model and the partial view scene point cloud is small, which makes them difficult to be registered.For example, \cite{prnet} mainly register two partial view point clouds and \cite{yang2020mapping} mainly register two full object meshes. As the IoU between point clouds in these cases is high, the methods work well. However, it fails in our cases when a partial view point cloud needs to be aligned with a full object mesh.  So far, the most similar method with us is Scan2CAD~\cite{scan2cad} that matches furniture CAD models with indoor RGB-D scan. Our task and method differ from theirs in four aspects: (a) our target scene is a single-view partial point cloud which is closer to a practical setting, while~\cite{scan2cad} focuses on a 3D reconstruction of an indoor scene, (b) our objects can be precisely matched to the scene targets, while~\cite{scan2cad} considers a CAD object set that can only be roughly matched with the scene targets, (c) both the objects and scenes have color information in our setting, while~\cite{scan2cad} only focuses on geometry matching and (d) we focus on table-top level object pose estimation while~\cite{scan2cad} focuses on larger scale in-door environment furniture alignment.

\subsection{Keypoint Features and Matching.}\label{subsec:relatedworks_matching}
Given two RGB images, conventional methods~\cite{strecha2011ldahash,lourencco2012srd,ke2004pca} use hand-crafted features such as SIFT~\cite{sift}, SURF~\cite{surf} and ORB~\cite{orb} for corresponding keypoints detection and matching. Recently, deep learning techniques have been applied to this long-standing area~\cite{superglue,superpoint} and show promising performances in both accuracy and efficiency. A similar trend also appears in the 3D area. Researchers proposed hand-crafted descriptors~\cite{fpfh,shot} in early years for point cloud registration. However, these methods are time-consuming and limited in performance. Point cloud based neural networks~\cite{pointnet,pointnetpp,ppfnet,fcgf} improved the performances and found a balance between efficiency and accuracy. Other researches~\cite{pointfusion,densefusion,ffb6d} extracted multi-modal features by fusion. The fusion of multi-modal information compensates for the limitations of any single-modal features and thus improves the overall performance.



\subsection{\Pose\ Estimation Metrics.}\label{subsec:relatedworks_metric}
So far, the most commonly used metrics in \pose\ estimation literature are ADD~\cite{add} and ADD-S~\cite{posecnn}. ADD measures the average point error between the estimated pose and the ground truth pose. It is intuitive but not applicable to rotational symmetric objects because of the problem of pose ambiguity. ADD-S metric is thus proposed to solve the problem. However, ADD-S is not a good measurement on pose error itself and can be problematic under some circumstances. Examples will be given in supplementary materials. ACPD and MCPD metrics~\cite{hua_evaluation_2016} are proposed which can handle objects with finite pose ambiguities in a unified manner. But, they still fail when objects have infinite ambiguous poses.
\section{Task Definitions}\label{sec:task}

\noindent\textbf{Point Cloud} is defined by a matrix $\mathcal{P} $
\begin{equation}
    \label{eqn:cpcd}
    \mathcal{P} = \left[ \begin{matrix}
    x_1 & y_1 & z_1& r_1 & g_1 & b_1 \\ 
    x_2 & y_2 & z_2& r_2 & g_2 & b_2 \\
    & & \cdots &\cdots\\
    x_n & y_n & z_n& r_n & g_n & b_n \\
    \end{matrix} \right]
\end{equation}

in which $x_i, y_i, z_i$ and $r_i, g_i,b_i$ represent the 3D coordinates and RGB values of the $i^{th}$ point respectively.

\noindent\textbf{Object \Pose} $T$ is an element of the special Euclidean group SE(3) that represents the object translation and rotation in the scene.
\begin{equation}
    \label{eqn:pose}
    T \in SE(3) 
\end{equation}


For the task of \task,
the input of the task is a tuple $I$.
\begin{equation}
    \label{eqn:input}
    I = (s, o)
\end{equation}
in which $s$ and $o$ represent the scene and object respectively. The scene $s$ is usually denoted by a colored point cloud which is captured by indoor RGBD cameras such as Intel RealSense or Lidar. The object $o$ is usually denoted by a triangle mesh model which can also be sampled and interpolated as a colored point cloud.

Unseen object 6D pose estimation algorithm $F$ is a function that maps the input tuple to a \pose, through which the object mesh can be transformed to its scene counterpart in the camera frame.
\begin{equation}
    F(I) = F(s,o) \rightarrow T
\end{equation}

The dataset $\mathcal{D}$ is composed of the training set $\mathcal{D}_{train}$ and test set $\mathcal{D}_{test}$.
\begin{equation}
\begin{array}{c}
    \mathcal{D} = \mathcal{D}_{train} \cup \mathcal{D}_{test}
    \\
    \mathcal{D}_{train} \cap  \mathcal{D}_{test} = \varnothing\\
    \end{array}
\end{equation}

Each element $d \in D$ is a tuple $d^i = (s^i, o^i, T^i)$ in which $T^i$ is the ground truth \pose. Assume $\mathcal{O}_{test}$ is the object set for the testing and $\mathcal{O}_{train}$ is the one for the training.
Previous algorithms focus on the problem when $\mathcal{O}_{train} \supseteq \mathcal{O}_{test}$. However, this \task\ task requires novel object in the test set. 
\begin{equation}
    \label{eqn:unseen}
    \begin{array}{c}
    \mathcal{O}_{train} = \left\{ o^i_{train}, i = 1, 2, \cdots, n_{train} \right\} \\
    \mathcal{O}_{test} = \left\{ o^i_{test}, i = 1, 2, \cdots, n_{test} \right\} \\
    \exists o \in \mathcal{O}_{test}, o \notin \mathcal{O}_{train} \\
    \end{array}
\end{equation}

Suppose $TP$, $M(TP, T, o)$ are the predicted pose and pose error metric. The task requires the algorithm $F$ to minimize the average pose error on the test set given the training set.
\begin{equation}
    \label{eqn:task}
    \argmin_{F}\frac{1}{n_{test}}\sum_{i=1}^{n_{test}}M(F(s_{test}^i, o_{test}^i), T_{test}^i, o_{test}^i) \left|  D_{train}\right.
\end{equation}

\begin{figure*}[t]
    \centering
    \includegraphics[width=0.90\linewidth]{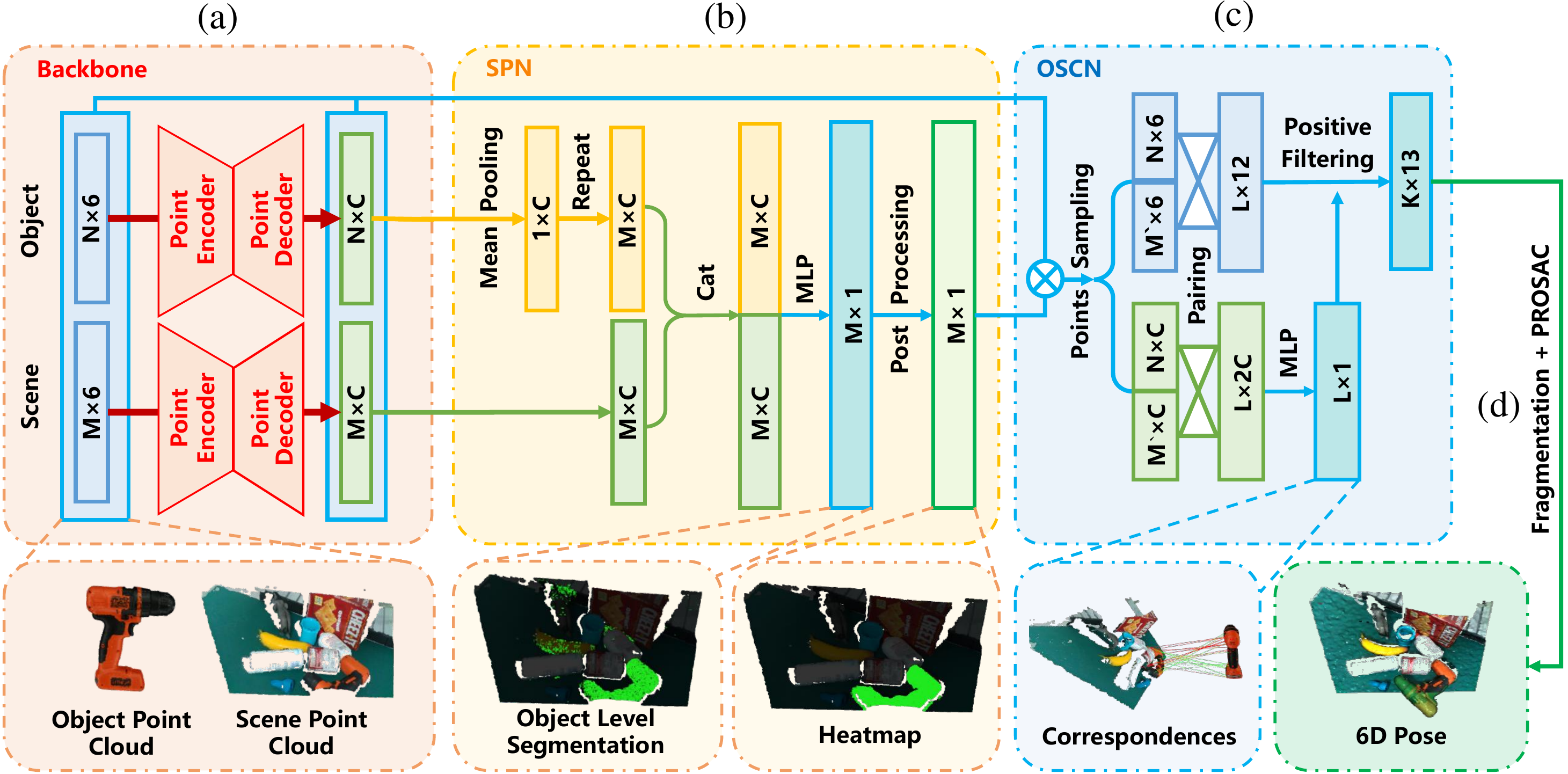}
    \caption{\textbf{Baseline Architecture}: The method could be divided into two stages. The first stage is an end-to-end neural network which detects 3D correspondences between the object and the scene. It is composed of three parts, i.e., backbone, \segmethodfull\ (\segmethod) and \correpmethodfull\ (\correpmethod). The second stage calculates the \pose\ from the 3D correspondences with PROSAC algorithm.}
    \label{fig:baseline}
\end{figure*}
\section{Method}\label{sec:baseline}

In this section, we provide a baseline solution for the \task\ task based on the architecture illustrated in Fig.~\ref{fig:baseline}. Given the point cloud of the object and scene, we start by extracting high dimensional features for the two inputs using a backbone network. Then \segmethodfull\ (\segmethod) proposes candidates of the target object in the scene using these features and we further obtains object ROIs(region of interest) with a manually selected threshold. After that, an \correpmethodfull\ (\correpmethod) learns the dense 3D correspondences between the object points and the scene points in each selected ROI region. Finally, we follow EPOS~\cite{epos} to estimate the object \pose\ from the 3D correspondences. 

\subsection{Backbone Network}\label{subsec:method_backbone}
Our framework starts with point cloud feature extraction. In our setting, As shown in Fig.~\ref{fig:baseline}a, given the object point cloud $\mathcal{P}_{obj}$ of size $N \times 6$ and scene point cloud $\mathcal{P}_{scene}$ of size $M \times 6$, the backbone network extracts point-wise high dimensional feature vectors $\mathcal{F}_{obj}$ and $\mathcal{F}_{scene}$ of shape $N \times C$ and $M \times C$ for each input respectively. These features are shared in the latter \segmethod\ and \correpmethod\ modules to segment the point cloud and find 3D correspondences.

\subsection{\segmethodfull (\segmethod)}
\label{sec:seg}
Finding correspondences in the whole scene is a difficult task. Inspired by ~\cite{graspness}, we introduce the attention mechanism by adding a point-wise segmentation network \segmethod. 
Given the features of the target object $\mathcal{F}_{obj}$ and that of the scene $\mathcal{F}_{scene}$, \segmethod\ is designed to achieve a point-wise segmentation of this object in the scene.
This helps the \correpmethod\ find correspondences as it can focus on a small area, which also saves the computational resources.

For the network structure, as shown in Fig.~\ref{fig:baseline}b, we first apply a mean pooling on the object features $\mathcal{F}_{obj}$ to obtain the object's global feature vector of shape $1 \times C$, which serves as a descriptor for the object. Then, we concatenate such feature vector with each point in the scene features $\mathcal{F}_{scene}$, resulting in a shape of $M \times 2C$. These concatenated features are fed into a multi-layer perceptron (MLP). The final output is a segmentation heatmap of shape $M \times 1$, denoting whether each point in the scene belongs to the object or not.

\begin{figure}[t]
    \centering
        \begin{subfigure}[b]{0.22\textwidth}
            \centering
            \includegraphics[width=\textwidth]{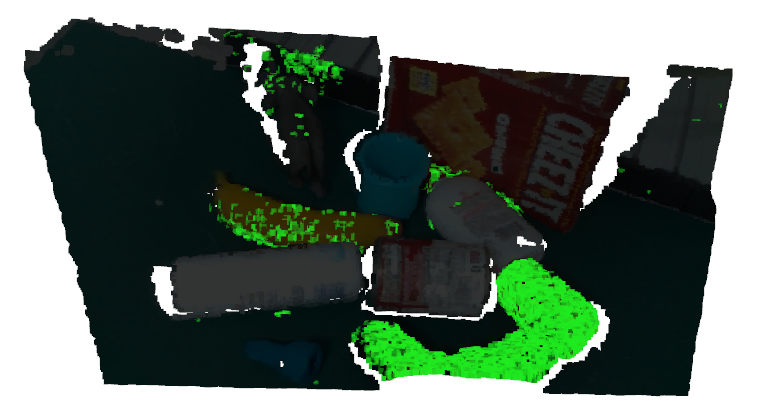}
            \caption{\segmethod 's result}
            \label{subfig:seg_raw}
        \end{subfigure}
        \hfill
        \begin{subfigure}[b]{0.22\textwidth}
            \centering
            \includegraphics[width=\textwidth]{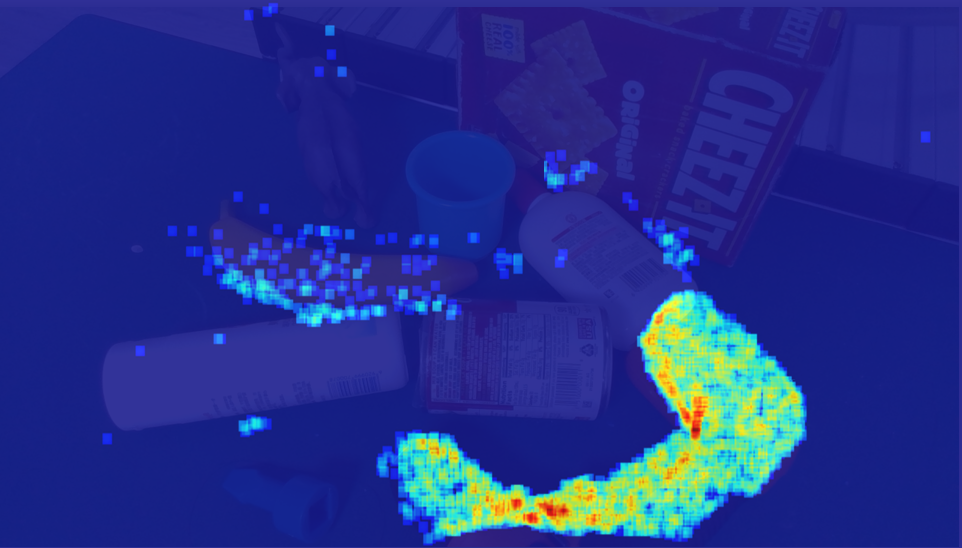}
            \caption{Heatmap}
            \label{subfig:heatmap}
        \end{subfigure}
        \hfill
        \begin{subfigure}[b]{0.22\textwidth}
            \centering
            \includegraphics[width=\textwidth]{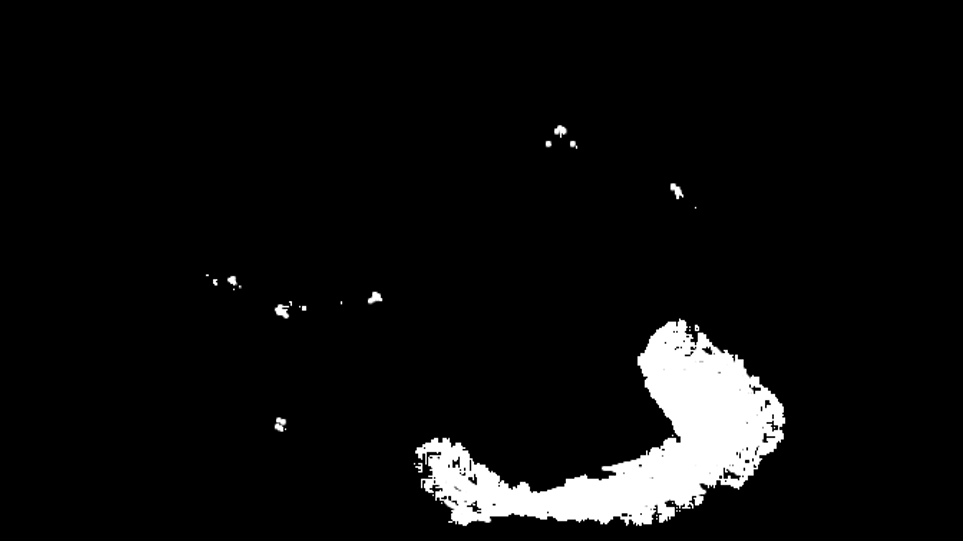}
            \caption{Otsu threshold}
            \label{subfig:seg_contour}
        \end{subfigure}
        \hfill
        \begin{subfigure}[b]{0.22\textwidth}
            \centering
            \includegraphics[width=\textwidth]{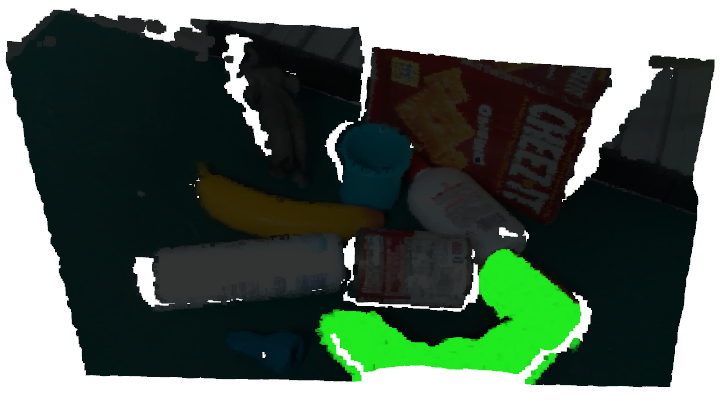}
            \caption{
            Segmentation}
            \label{subfig:seg_roi}
        \end{subfigure}
    \caption{(a) is the output of \segmethod. (b) is the heatmap after Gaussian smoothing. (c) is the result of Otsu method~\cite{otsu}. (d) is the final segmentation result.}
    \label{fig:seg}
\end{figure}
To cover as many points on the object as possible and eliminate noise, we conduct a post-processing to refine the segmentation heatmap. As shown in Fig.~\ref{fig:seg}, we first project the 3D heatmap to the 2D scene image and apply a Gaussian smoothing to the 2D heatmap. This makes the discretized heatmap more continuous.
Then, to remove outliers, we binarize the heatmap by Otsu's method~\cite{otsu} and select the connected components larger than a threshold as the final segmentation results which is shown in Fig.~\ref{subfig:seg_roi}.

\subsection{\correpmethodfull\ (\correpmethod)}
After obtaining the target object segmentation candidates, the \correpmethod\ module then finds 3D correspondences between each segmented scene ROI and the object. This module follows different strategy during training and testing, and we first introduce the testing stage.

During testing, the object level segmentation results are used to segment the target object candidates in the scene. For simplicity, we consider the case of only one target object candidate, while the case of multiple targets can be processed batch-wise similarly. Given the target segmentation, we crop both the input scene point cloud and its features and obtain $\mathcal{P}_{seg}$ and $\mathcal{F}_{seg}$, which have a shape of $M' \times 6$ and $M' \times C$. Then, for each point on the object and the segment scene, we concatenate their features and construct dense pair-wise feature vectors with a shape of $(M' \times N) \times 2C$, where $(M' \times N)$ denotes the amount of object-scene point pairs. To save computation resources, we randomly sample $L$ pairs and feed them into an MLP, which estimates $L \times 1$ scores ranging from 0 to 1 to denote the confidence of input pairs' correspondences. 

Among the $L$ point pairs, we select those with confidence score larger than $0.8$, resulting in $K$ corresponding point pairs. The point cloud of these corresponding points as well as the correspondence's scores are used for the final \pose\ computation, which is detailed in Sec.~\ref{subsec:6pc}.


During training, the segmentation results from \segmethod\ is not used. Instead, we uniformly sample $k_1$ pairs of matching points between the object and the scene using the ground truth \pose\ and randomly sample $k_2$ pairs of non-matching points as negative samples. The ratio $r_k = k_1 : k_2$ is a fixed hyper parameter to ensure a balanced training set.  


\subsection{6D Pose Computation}
\label{subsec:6pc}
As discussed in Section~\ref{sec:relatedworks}, the traditional way to obtain \pose\ from 3D correspondences is least square fitting with RANSAC~\cite{ransac}. However, such method would fail when objects have keypoint ambiguity which is usually caused by symmetry. In this paper, we adopt the 6D pose fitting module proposed in EPOS~\cite{epos} for the final 6D pose computation. It adopts the PROSAC algorithm~\cite{prosac} instead of RANSAC~\cite{ransac} to calculate the final 6D pose. This algorithm is a locally optimized RANSAC that firstly focuses on correspondences with higher confidence and progressively turns to uniform sampling. For more details, we refer readers to the original paper of EPOS~\cite{epos}.





\subsection{Loss}\label{subsec:method_loss}
The backbone, \correpmethod\ and \segmethod\ modules are trained simultaneously with multi-task loss:
\begin{equation}
\label{eqn:loss}
L = (1-\lambda) L_{seg} + \lambda \ L_{cor}, (0 < \lambda < 1),
\end{equation}
where $L_{seg}$ and $L_{cor}$ are both binary cross entropy loss for target classification in \segmethod\ and correspondence classification in \correpmethod\ respectively. 
\section{Implementation Details}\label{sec:implementation}

\paragraph{Dataset.} To construct a meaningful benchmark, it requires a variational training set so that the networks can learn representations general enough and a representative test set that is close to the real-world setting. GraspNet-1Billion~\cite{graspnet}, originally proposed for the problem of robotic grasping, satisfies most of our requirements.  It contains 40 objects and 100 scenes for training, 76 objects and 90 scenes for testing. Its test set is further divided into 3 subset, namely seen object set, similar object set and novel object set, where each set contains 30 scenes consisted of 28 seen objects, 22 unseen but similar objects and 26 totally novel objects respectively. Thus, we build our benchmark upon~\cite{graspnet}. The only problem is that its training set only contains 40 objects, which may be too few for the network to learn model-agnostic geometry correspondence features. Thus, we generate extra synthetic training data with BlenderProc~\cite{blenderproc} simulator. The object mesh models come from the Google Scanned Object dataset~\cite{google1000}, which consists of over 1000 real-world objects. In total, there are $1070$ objects and $1500$ scenes ($1400$ synthetic scenes and $100$ real scenes from Graspnet-1Billion ) in our training set and $76$ objects and $90$ scenes in the real data test set, in which $48$ objects are unseen during training. 

To verify the effectiveness of our method, we also conduct pose estimation experiments on the YCB-Video~\cite{posecnn} without any retraining or fine-tuning and compare the results of different algorithms.

\paragraph{Neural Network and Training.}
For the backbone network, we select ResUNet14 built on MinkowskiEngine~\cite{minkowski} which has great performance in processing the point cloud. It can also be replaced by other point cloud networks such as PointNet~\cite{pointnet} and PointNet++~\cite{pointnetpp}. $M$ and $N$ are the points number of the scene and object's point cloud. $C$ is set to $512$. $L$ is set to $102400$ during inference. $k_1$ and $k_2$ are set to $100$ and $600$ during training. All the MLPs are implemented using full connected layers with residual blocks. The structure is illustrated in the supplementary materials.
The $\lambda$ value in the loss layer is set to $0.6$. To reduce the size of the neural network, the backbone for object branch and scene branch share the same structure and weights. 
Our model is implemented with PyTorch and is trained and tested on a server with 8 NVIDIA RTX 3090 GPUs. The backbone, \segmethod\ and \correpmethod\ are trained simultaneously by minimizing the loss described in Section~\ref{subsec:method_loss} with Adam optimizer~\cite{adam} for 10 epochs. The batch size is 3 and the learning rate is set to $10^{-3}$. 

\paragraph{Data Augmentation.}
We conduct heavy data augmentation to avoid over-fitting. 
Before being fed into the network, the point clouds are voxel-downsampled with the voxel size of $0.002m$. The scenes' point clouds are augmented on-the-fly by random rotation around Z-axis in $\left[-\pi , \pi\right)$, while the object's point clouds are augmented by rotation around a random axis with a random degree in $\left[-\pi , \pi\right)$.

\section{Experiments}\label{sec:experiments}
We conduct extensive experiments to verify the effectiveness and efficiency of our proposed method. 

\subsection{Metric}\label{subsec:implementation_metric}
The biggest challenge of \pose\ evaluation metric is pose ambiguity~\cite{manhardt_explaining_2019}.
As discussed in Section~\ref{subsec:relatedworks_metric}, the most commonly used metrics are ADD~\cite{add} for objects with no pose ambiguity and ADD-S~\cite{posecnn} for object with pose ambiguity. But the two metrics cannot be compared because ADD-S is always numerically smaller than ADD~\cite{add}. In other words, the pose estimation result for symmetric objects cannot be compared with asymmetric objects. ACPD and MCPD~\cite{hua_evaluation_2016} are proposed to solve this problem which comprehensively evaluate all reasonable ground truth poses. But neither the definition itself nor the implementation~\cite{boptoolkit} is able to handle objects with infinite pose ambiguities. 
We propose a new metric named \textbf{I}nfimum of \textbf{ADD}(\metric). \metric\ extends ACPD when there are infinite pose ambiguities. 

The previous metrics are given in Equation~\ref{eqn:metrics}.
\begin{equation}
    \label{eqn:metrics}
    \begin{aligned}
        \mbox{ADD} &=  \frac{1}{m} \sum_{v \in \mathcal{V}}\left\| \left( Rv + T\right) - \left(R^*v+T^*\right) \right\|,\\
        \mbox{ADD-S} &=  \frac{1}{m} \sum_{v_1 \in \mathcal{V}}\min_{v_2 \in \mathcal{V}}\left\| \left( Rv_1 + T\right) - \left(R^*v_2+T^*\right) \right\|,\\
        \mbox{ACPD} &= \frac{1}{m} \sum_{v \in \mathcal{V}}\min_{R^* \in \mathcal{R}^*, T^* \in \mathcal{T}^*}\left\| \left( Rv + T\right) - \left(R^*v+T^*\right) \right\|,\\
    \end{aligned}
\end{equation}

where $\mathcal{V}$, $v$, $R$, and $T$ are the vertex points set of the object, vertex point, rotational matrix, translation respectively. $R^*$, $T^*$, $\mathcal{R}^*$ and $\mathcal{T}^*$ denote the ground truth rotational matrix, translation and the set of ground truth rotational matrices and translations.
\begin{figure}[t]
    \centering
    \includegraphics[width=0.90\linewidth]{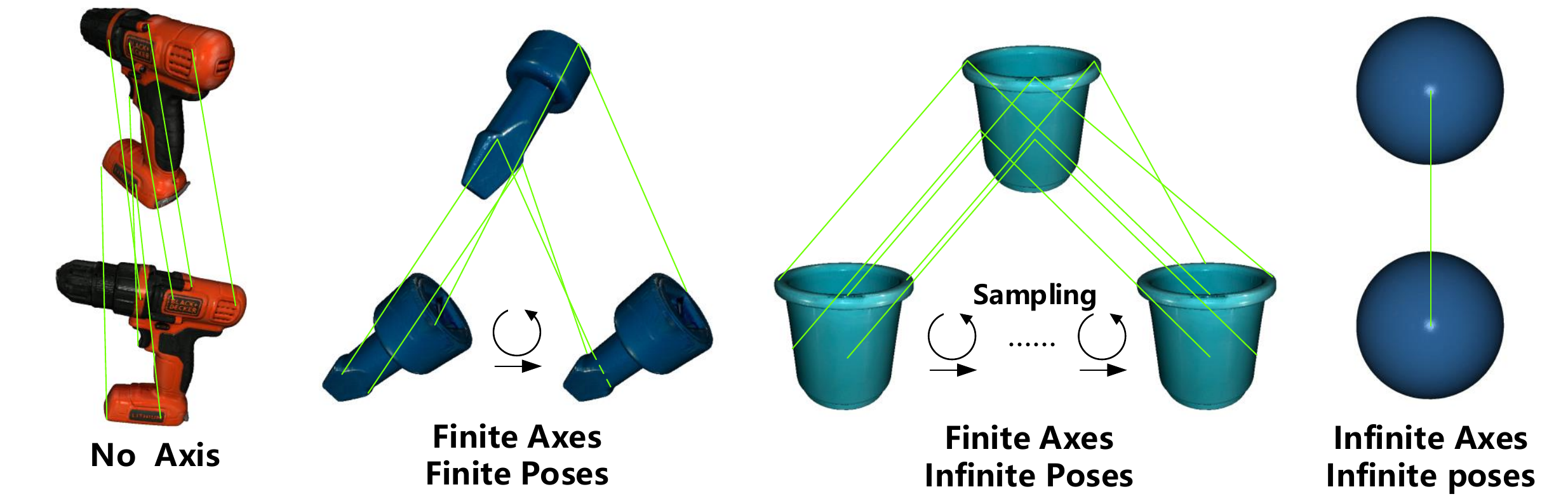}
    \caption{The illustration of \metric. In the first two cases, \metric\ equals to ACPD and can be calculated by traversing all the poses and find the minimum ADD. In the third case where the object has at least one rotational axis that has infinite pose ambiguities, we estimates the infimum of ADD by uniform sampling. In the last case where the object has infinite rotational axes, \metric\ is the distance between the ground truth center and estimated center.}
    \label{fig:metric}
\end{figure}
The definition of \metric\ is given in Equation~\ref{eqn:badd}.
\begin{equation}
    \label{eqn:badd}
    \mbox{\metric} = \frac{1}{m} \sum_{v \in \mathcal{V}}\inf\limits_{R^* \in \mathcal{R}^*, T^* \in \mathcal{T}^*}\left\| \left( Rv + T\right) - \left(R^*v+T^*\right) \right\|
\end{equation}

Using this metric, both the pose for symmetric and asymmetric objects can be evaluated in the same way even if there are infinite pose ambiguities. To further discuss the implementation of this metric, we firstly discuss where the pose ambiguity comes from.

Pose ambiguity occurs only when the object has a rotational symmetry axis. Mirror symmetry brings problems of keypoint ambiguity for \pose\ estimation algorithms. But it leads to no pose ambiguity in evaluation. As shown in Fig.~\ref{fig:metric}, there are totally four cases.
\begin{enumerate}
    \item Object has no rotational axis.
    \item Object has finite rotational axes and each rotational axis has finite equivalent poses.
    \item Object has finite rotational axes and at least one rotational axis has infinite equivalent poses. 
    \item Object has infinite rotational axes.
\end{enumerate}

For the first case, \metric\ equals to ADD and ACPD. For the second case, \metric\ equals to ACPD. For the third case, it is hard to find an analytical solution. We sample $n$ angles around the axis with infinite pose ambiguities in our implementation. The number of $n$ is a trade-off between precision and efficiency. Although this is a numerical solution, it doesn't destroy the overall science as ADD itself is a numerical solution that samples points from a mesh model. For the last case, although we can still take the numerical solution, the sampling on two dimensions, i.e. the axis sampling and the rotation angle sampling, results in a huge cost for computation. Fortunately, the only object that has infinite rotational axes is a texture-less sphere. For this kind of objects, \metric\ equals to the center distance between the target pose and the estimated pose.

\subsection{Experiment Results}\label{subsec:experiments_results}
\subsubsection{Visualization of Extracted Features}\label{subsubsec:experiments_results_features}
\begin{figure}[t]
    \centering
        \hfill
        \begin{subfigure}[b]{0.25\columnwidth}
            \centering
            \includegraphics[width=0.5\columnwidth]{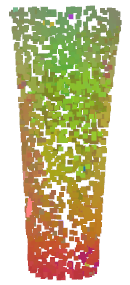}
            \caption{Object}
            \label{fig:tsne_l}
        \end{subfigure}
        \hfill
        \begin{subfigure}[b]{0.6\columnwidth}
            \centering
            \includegraphics[width=0.8\columnwidth]{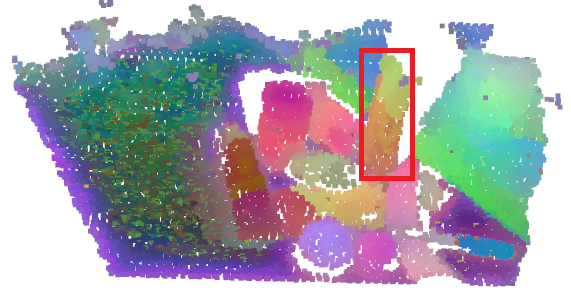}
            \label{fig:tsne_r}
            \caption{Scene}
        \end{subfigure}
        \hfill
    \caption{t-SNE \cite{tsne} visualization result of encoded features presented in RGB.}
    \label{fig:tsne}
    \vspace{-0.2in}
\end{figure}

We reduce the dimensions of extracted features to 3 and colorize the point cloud by encoding the RGB channel with these 3D features. As shown in the visualization result in Fig.~\ref{fig:tsne}, both the features among different objects and those among different parts within an object are clearly distinguishable.




\begin{table*}[t]
    \caption{Quantitative results of different methods. g.t., w.o., G. and YCB-V are short for ground truth, without, GraspNet-1Billion~\cite{graspnet} and YCB-Video~\cite{posecnn} respectively. The number is the area under curve(AUC) score of each methed using each metric. The upper bound of AUC is set to $0.5 \times $ diagonal of the target object.}
    \begin{center}
    
        \begin{tabular}{cccccc}
            \toprule
            \multirow{2}[4]*{Method} & \multirow{2}[4]*{Metric} &  \multicolumn{4}{c}{Dataset} \\
        \cmidrule{3-6}          &    & \ G. Seen\  &\ G. Similar\ &\ G. Novel\ &\ YCB-V \\
            \midrule
            SuperGlue~\cite{superglue} & ADD    & 10.5     & 8.5     & 5.5     & 7.6 \\
            +     & ADD-S                       & 13.5     & 11.8     & 7.6     & 12.5 \\
            RANSAC~\cite{ransac} & \metric     & 10.6     & 8.8     & 6.7     & 8.1 \\
            \midrule
            ICP~\cite{icp}   & ADD     & 2.1     & 4.5     & 5.6     & 6.7 \\
            +     & ADD-S     & 8.7     & 11.6     & 12.7     & 13.5 \\
            DBSCAN~\cite{dbscan} & \metric     & 2.2     & 5.1     & 5.9     & 7.1 \\
            \midrule
            FCGF~\cite{fcgf}  & ADD       & 13.5     & 10.4     & 11.5     & 6.6 \\
                        +     & ADD-S      & 57.9     & 51.0     & 44.6     & 49.3\\
            \segmethod. & \metric        & 15.0     & 12.6     & 13.1     &7.7      \\

            \midrule
            FCGF~\cite{fcgf}  & ADD       & 1.5     & 1.4     & 2.3     & 1.1 \\
        w.o.                  & ADD-S    & 14.7     & 15.4     & 15.3     & 11.6\\
        \segmethod & \metric               & 1.7     & 1.8     & 2.7     & 1.5 \\
            \midrule
            Ours                & ADD       & 3.0     & 2.8     & 3.8     & 2.1 \\
            w.o.                & ADD-S     & 20.0     & 20.2     & 22.8     & 16.8 \\
            \segmethod & \metric            & 3.1     & 3.3     & 4.4     & 2.4 \\
            \midrule
                  & ADD       & \textbf{33.8} & \textbf{25.3}     & \textbf{21.2}     & \textbf{17.8} \\
            Ours  & ADD-S      & \textbf{65.6} & \textbf{58.1}     & \textbf{50.3}     & \textbf{55.6} \\
                  & \metric     & \textbf{36.3} & \textbf{29.2}     & \textbf{23.7}     & \textbf{19.0} \\
            \bottomrule
        \end{tabular}%
    \end{center}
    \vspace{-0.2in}

    \label{tab:eval} 
\end{table*}%

\subsubsection{Qualitative and Quantitative Results on \Pose}\label{subsubsec:experiments_results_poseresult}
As discussed in Section~\ref{sec:relatedworks}, no previous work proposes solution to this new task. We implement several baseline methods based on both deep-learning and conventional algorithms. These baselines include point cloud clustering~\cite{dbscan} + ICP registration~\cite{icp} , SuperGlue~\cite{superglue} + RANSAC~\cite{ransac} + least square fitting and FCGF~\cite{fcgf} + \segmethod. 

\begin{figure}[t]
    \centering
    \includegraphics[width=1\linewidth]{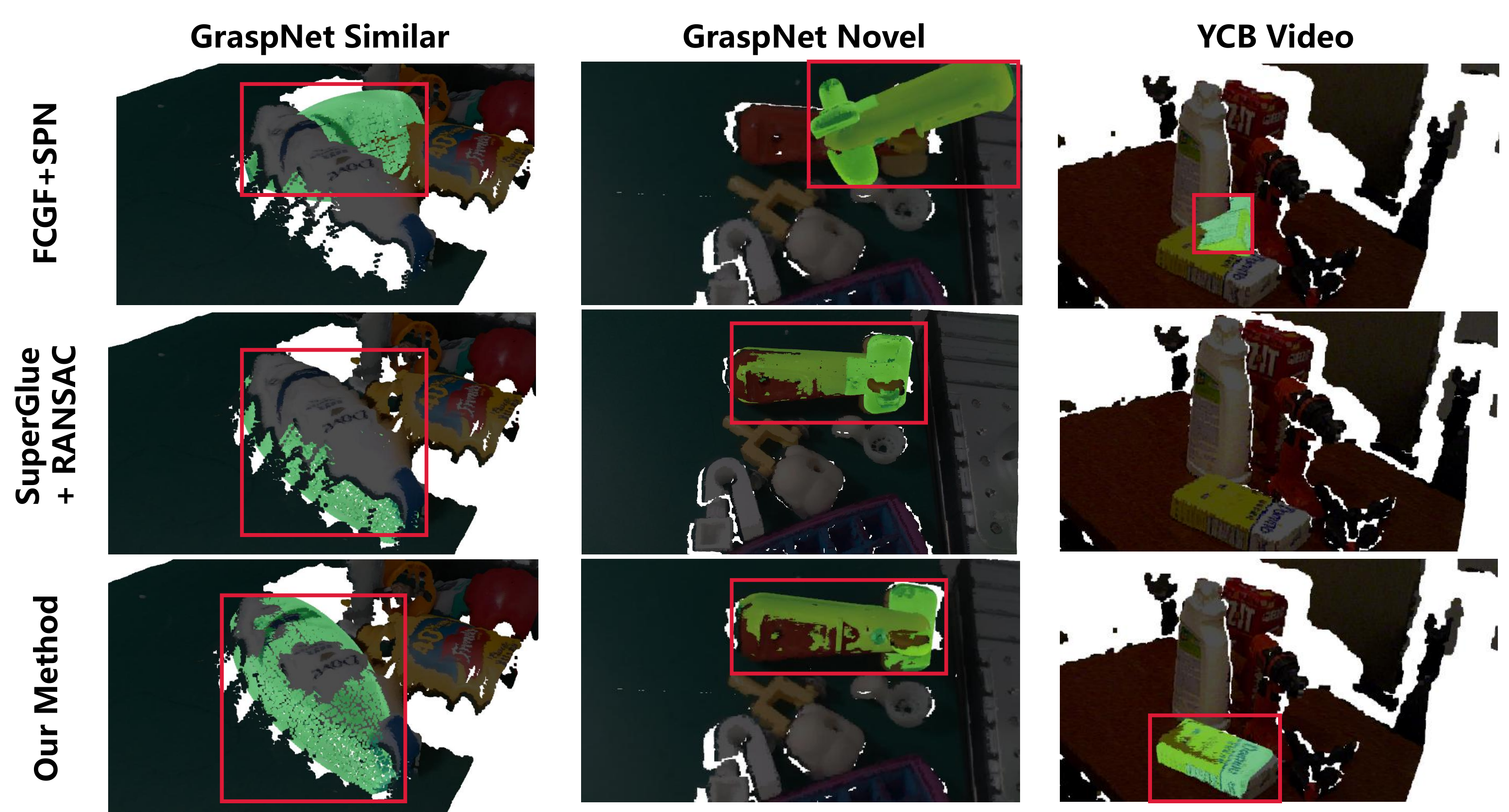}
    \caption{The qualitative result of pose estimation on one object in the scene. The selected object is painted green and identified by a red bounding box. We can see that FCGF + SPN cannot generate satisfactory results. SuperGlue + RANSAC fails to estimate the pose for the image in YCB-Video dataset due to the target is too small. Our method performs well across different scenes.
    }
    \label{fig:result}
\end{figure}

\begin{figure}[t]
    \centering
    \includegraphics[width=\linewidth]{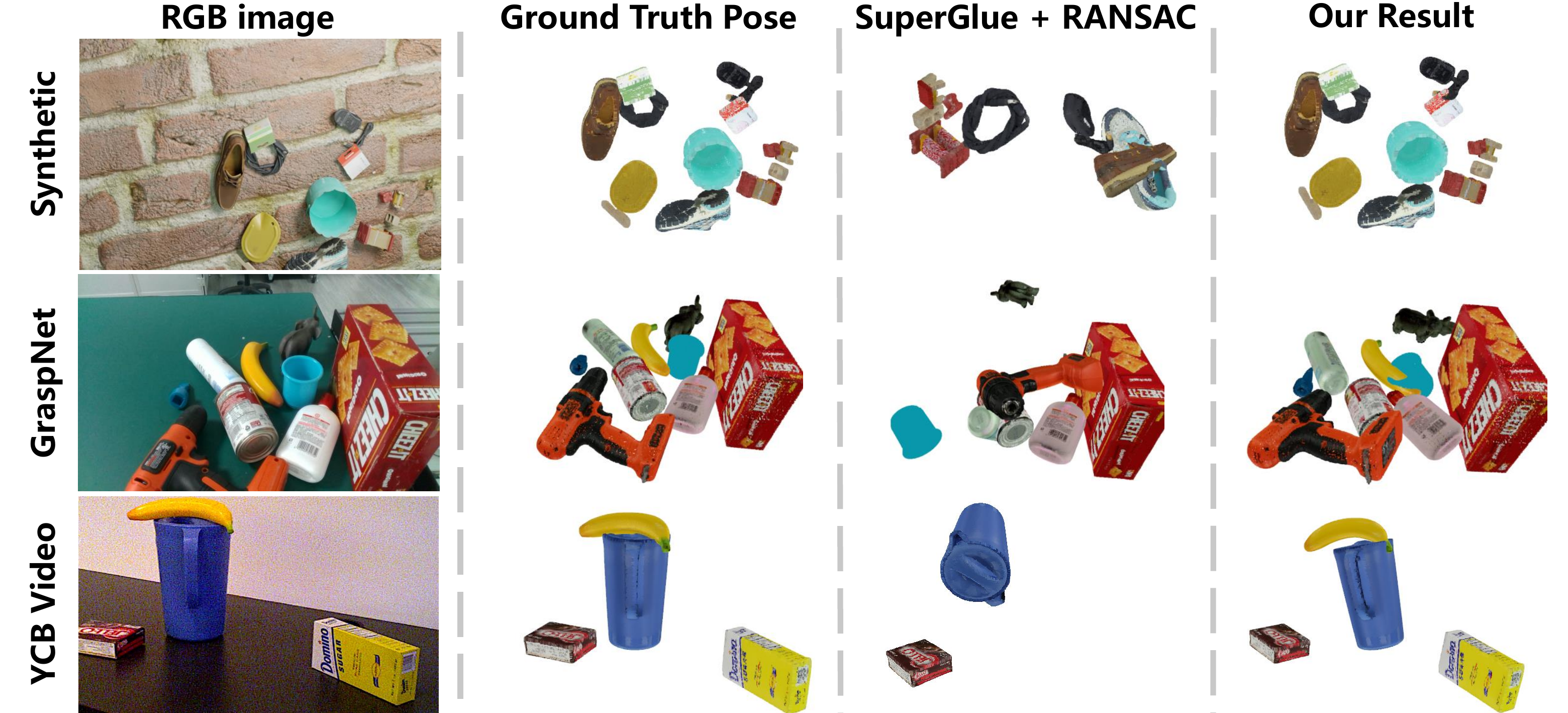}
    \caption{The qualitative result of pose estimation on all objects in the scene.}
    \label{fig:scene}
\end{figure}

The quantitative results using ADD, ADD-S and IADD metrics are reported in Table~\ref{tab:eval}. We can see that our method outperforms other baselines by a large margin in all the test subset. From the AUC scores of different metrics across different test subset, we can see that IADD is more numerical stable than ADD-S. For example, both ADD and IADD reports a lower scores for our method on YCB-V than on G.Novel subset, while ADD-S reports a better performances. We also show the qualitative results of object \pose\ generated by different methods in Fig.~\ref{fig:result} and Fig.~\ref{fig:scene}.
Our method is more robust compared with other baselines. 


\subsubsection{3D Correspondences}\label{subsubsec:experiments_results_correspondence}
\begin{figure}[t]
    \centering
    \hfill
    \begin{subfigure}[b]{0.30\columnwidth}
        \centering
        \includegraphics[width=\textwidth]{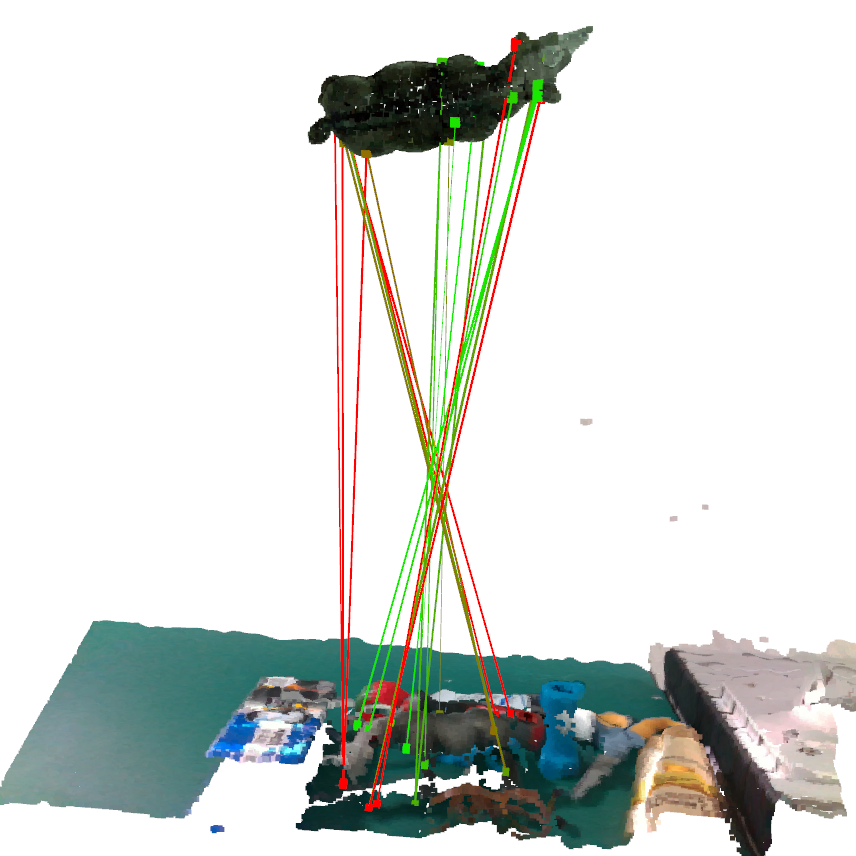}
        \caption{\correpmethod's result}
        \label{subfig:corresp_ours}
    \end{subfigure}
    \hfill
    \begin{subfigure}[b]{0.30\columnwidth}
        \centering
        \includegraphics[width=\textwidth]{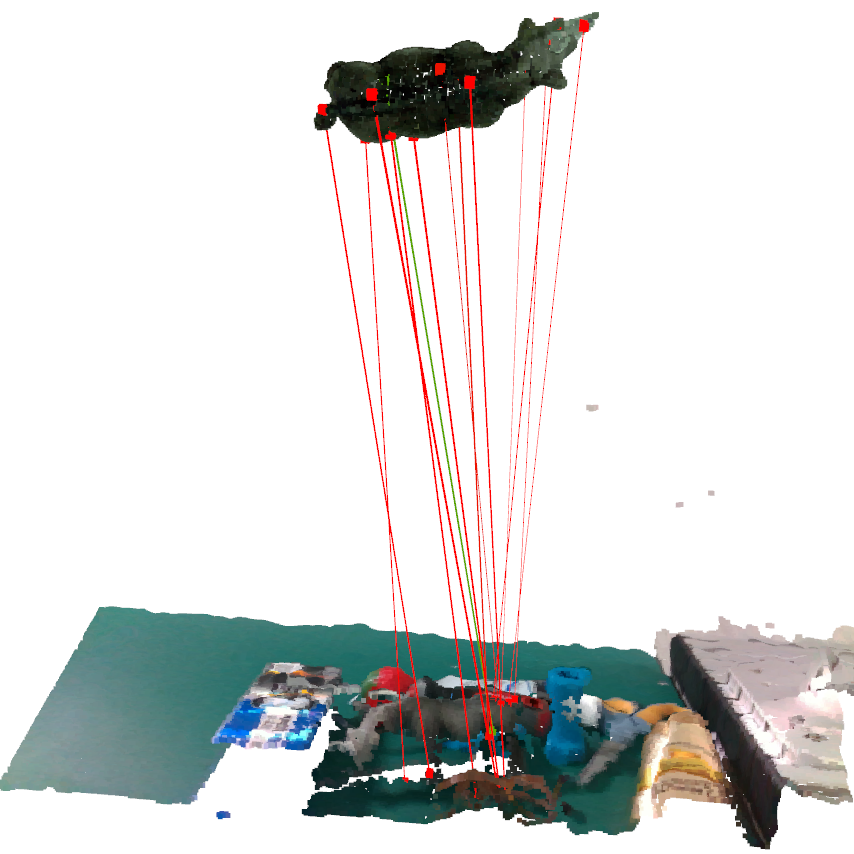}
        \caption{FCGF~\cite{fcgf} + \segmethod}
        \label{subfig:corresp_fcgf}
    \end{subfigure}
    \hfill
    \begin{subfigure}[b]{0.30\columnwidth}
        \centering
        \includegraphics[width=\textwidth]{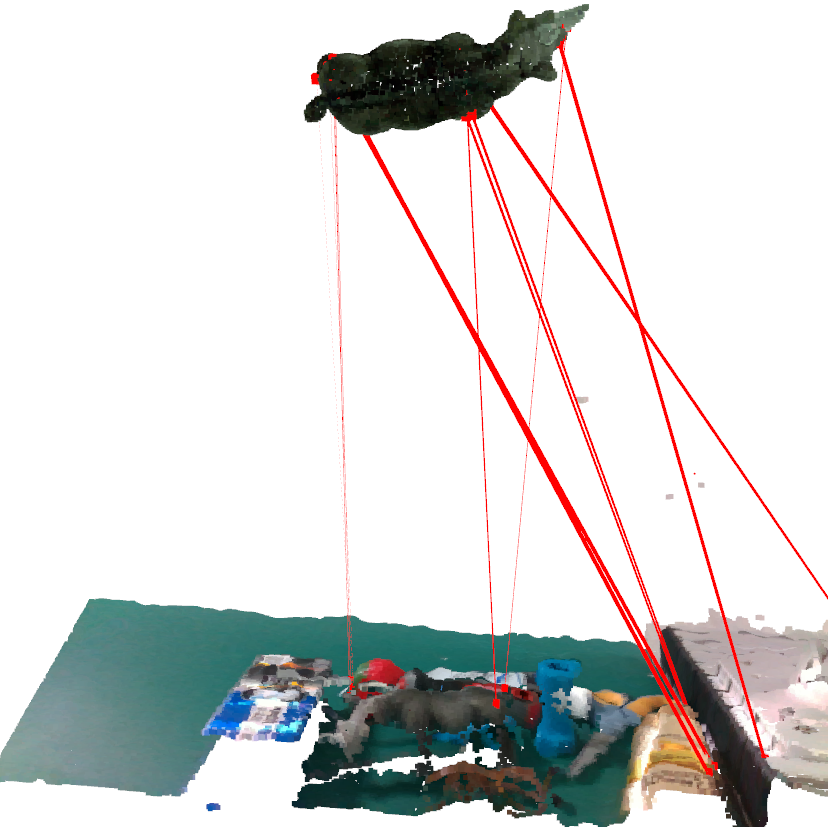}
        \caption{FCGF~\cite{fcgf}}
        \label{subfig:corresp_fcgf_wo_seg}
    \end{subfigure}
    \hfill
    \caption{Visualization of different correspondences establishing algorithms. The color of the lines represent the error of correspondences. The greener line has smaller error while the redder line has larger error.}
    \label{fig:correspondence}
\end{figure}

The 3D keypoints correspondences matching result is shown in Fig.~\ref{fig:correspondence}. The FCGF~\cite{fcgf} is originally designed for point cloud registration. However, when it comes to this task, its performances are far from satisfactory, especially for small objects. As a result, it is not suitable for the \task\ task. A potential reason is that the scene point cloud is only partially observable and has noises. In contrast, our method is more powerful in finding the correspondences against noise, which is beneficial for obtaining the object \pose. 

\subsubsection{Efficiency Comparison}\label{subsubsec:experiments_results_efficiency}
To evaluate the efficiency of different methods, we conduct experiments using a computer with single NVIDIA RTX 3070 GPU with 16 cores CPU. The average inference time cost is recorded in Table~\ref{tab:speed}. Not only does our method outperforms other baselines, but it also costs less time.
\begin{table}[htbp]
  \caption{Average inference time cost of different methods.}
  \begin{center}
    \begin{tabular}{c|p{25mm}}
    \toprule
    Method & {\small Average Inference Time(s)} \\
    \midrule
    ICP~\cite{icp} + DBSCAN~\cite{dbscan}   & $\sim42$ \\
    SuperGlue~\cite{superglue} + RANSAC~\cite{ransac} & $\sim15$\\
    FCGF~\cite{fcgf} + \segmethod&  $\sim\ $\textbf{3}\\
    ours  & $\sim4$ \\
    \bottomrule
    \end{tabular}%
  \end{center}
  
  \label{tab:speed}
\end{table}%

\subsubsection{Effectiveness of SPN}\label{subsubsec:experiments_results_ablation}
We also conduct an ablation study to verify the importance of \segmethod. As shown in Table~\ref{tab:eval}, when \segmethod\ is removed, the score of our method drops dramatically. By introducing the attention mechanism of \segmethod, it is much easier to find corresponding points between the object and scene. 

\section{Conclusion}\label{sec:conclusion}
In this paper, we propose a new task named unseen object 6D pose estimation. A large-scale dataset that can fulfill this task is collected and generated. A two-stage baseline method for the task is developed to detect unseen objects' 6D pose by finding 3D correspondences. A new metric named \metric\ is also presented to evaluate the pose estimation result for objects with all kinds of pose ambiguity in the same way. The experiments show that our method greatly outperforms several baselines in both accuracy and efficiency. Our benchmark, evaluation codes and baseline methods will be made publicly available to facilitate future research.


{\small
\bibliographystyle{ieee_fullname}
\bibliography{egbib}
}
\newpage
\section{Appendix}

\subsection{Detailed Discussion on 6D Pose Estimation Metrics}\label{sec:metric}

As discussed in Section 6.1 in the main paper, the metric $M$ of 6D pose estimation gives a quantitative evaluation on the error given the object mesh, ground truth pose and predicted pose. Note that the object ground truth pose may have more than one value since for some object, there are more than one correct pose. For example, any rotation of a texture-less sphere is correct in pose estimation. As a result, It has infinite ground truth poses. Thus, we denote the ground truth object poses as a set $\mathcal{P}_{gt}$.
Given the object mesh $o$, ground truth pose set $\mathcal{P}_{gt}$ and predicted pose $P_{pred}$, we have:
\begin{equation}
    \label{eqn:def}
    \begin{split}
    M: (o, P_{pred}&, \mathcal{P}_{gt})  \rightarrow m,\\
    m  &\in \mathbb{R}.\\
    \end{split}
\end{equation}

There are totally four metrics that are mentioned, i.e., ADD~\cite{add}, ADD-S~\cite{posecnn}, ACPD~\cite{hua_evaluation_2016} and IADD. In this part, we will only analyze ADD and ADD-S.

According to the general definition of metric function~\cite{chirikjian2015partial}, a metric should meet the following three requirements. 

\begin{equation}
    \label{eqn:r1}
    d(H_1, H_2) = 0 \Longleftrightarrow H_1 = H_2\quad\mbox{(Zero Value)}
\end{equation}

\begin{equation}
    \label{eqn:r2}
    d(H_1, H_2) = d(H_2, H_1)\quad\mbox{(Commutability)}
\end{equation}

\begin{equation}
    \label{eqn:r3}
    d(H_1, H_2) + d(H_2, H_3) \ge d(H_1, H_3)\quad\mbox{(Triangle Inequality)}
\end{equation}
in which $d$ and $H$ are the metric function and value to be evaluated respectively. 

In the following part, we will prove that neither ADD nor ADD-S is a good metric that can evaluate the 6D pose for objects both with and without pose ambiguity in the same way. We will give the proof on the basis of the three requirements above.

\subsection{Metrics without Pose
Ambiguity}\label{subsec:metric_no_ambiguity}
 
When there is no pose ambiguity, the set of ground truth pose set $\mathcal{P}_{gt}$ degenerates into a pose $P_{gt}$. The requirements described in Equation~\ref{eqn:r1},~\ref{eqn:r2} and~\ref{eqn:r3} for 6D pose estimation metric $M$ should be modified into Equation~\ref{eqn:general_metric_1},~\ref{eqn:general_metric_2} and~\ref{eqn:general_metric_3}.
\begin{equation}
    \label{eqn:general_metric_1}
        M(o, P_{gt}, P_{pred}) = 0  \Longleftrightarrow P_{gt} = P_{pred} \quad \mbox{(Zero Value)}
\end{equation}
\begin{equation}
    \label{eqn:general_metric_2}
        M(o, P_{gt}, P_{pred})  = M(o, P_{pred}, P_{gt}) \quad \mbox{(Commutability)}
\end{equation}
\begin{equation}    
    \label{eqn:general_metric_3}
    \begin{split}
        M(o, P_{gt}, P_{pred}^1) + M(o, P_{pred}^2, P_{gt}) 
        \ge M(o, P_{pred}^1, P_{pred}^2) \\ \mbox{(Triangle Inequality)}
    \end{split}
\end{equation}

\subsubsection{ADD.}
ADD meets the requirements above. The proof is given below.

\begin{proof}
\emph{Requirement 1.}
If $\mbox{ADD}(o, P_{gt}, P_{pred}) = 0 $, each point of the object with the target pose and the predicted pose aligns exactly, it is obvious that $P_{gt} = P_{pred}$.
\end{proof}

\begin{proof}
\emph{Requirement 2.}
\begin{equation}
\begin{split}
\mbox{ADD}(o, P_{gt}, P_{pred})   \\
=\frac{1}{m} \sum_{v \in \mathcal{V}} \left\| \left( R_{pred}v + T_{pred}\right) - \left(R_{gt}v+T_{gt}\right) \right\| 
\end{split}
\end{equation}

\begin{equation}
 =  \frac{1}{m} \sum_{v \in \mathcal{V}} \left\| \left( R_{gt}v + T_{gt}\right) - \left(R_{pred}v+T_{pred}\right) \right\| 
\end{equation}

\begin{equation}
= \mbox{ADD}(o, P_{pred}, P_{gt}) 
\end{equation}
\end{proof}

\begin{proof}
\emph{Requirement 3.}
As Euclidean distance between points in 3D space is already a metric function which meets the requirements described in Equation~\ref{eqn:r1},~\ref{eqn:r2} and~\ref{eqn:r3}. For $\forall v \in \mathcal{V}$, 
\begin{equation}
\begin{split}
\left\| \left( R_{pred}^{1}v + T_{pred}^{1}\right) - \left(R_{gt}v+T_{gt}\right) \right\| \\ + \left\| \left(R_{gt}v + T_{gt}\right) - \left(R_{pred}^{2}v+T_{pred}^{2}\right) \right\|\\
\ge 
\left\| \left( R_{pred}^{2}v + T_{pred}^{2}\right) - \left(R_{pred}^{1}v+T_{pred}^{1}\right) \right\|
\end{split}
\end{equation}

As a result,

\begin{equation}
\begin{split}
\frac{1}{m} \sum_{v \in \mathcal{V}} ( \left\| \left( R_{pred}^{1}v + T_{pred}^{1} \right) - \left( R_{gt}v+T_{gt} \right) \right\| +\\
\left\| \left( R_{gt}v + T_{gt} \right) - \left( R_{pred}^{2}v+T_{pred}^{2} \right) \right\| ) \\ \ge \left\| \left( R_{pred}^{2}v + T_{pred}^{2} \right) - \left( R_{pred}^{1}v+T_{pred}^{1} \right) \right\| \\ \Longrightarrow \mbox{ADD}(o, P_{gt}, P_{pred}^1) + \mbox{ADD}(o, P_{pred}^2, P_{gt}) \\ \ge \mbox{ADD}(o, P_{pred}^1, P_{pred}^2) 
\end{split}
\end{equation}

\end{proof}

\subsubsection{ADD-S.}
However, ADD-S doesn't meet at least two requirements for which the counter-examples are as follows. As a result, ADD-S is not a reasonable metric in this situation.
\begin{proof}
\emph{Violation of requirement 1.}
\begin{figure}[htbp]
    \centering
    \includegraphics[width=0.8\linewidth]{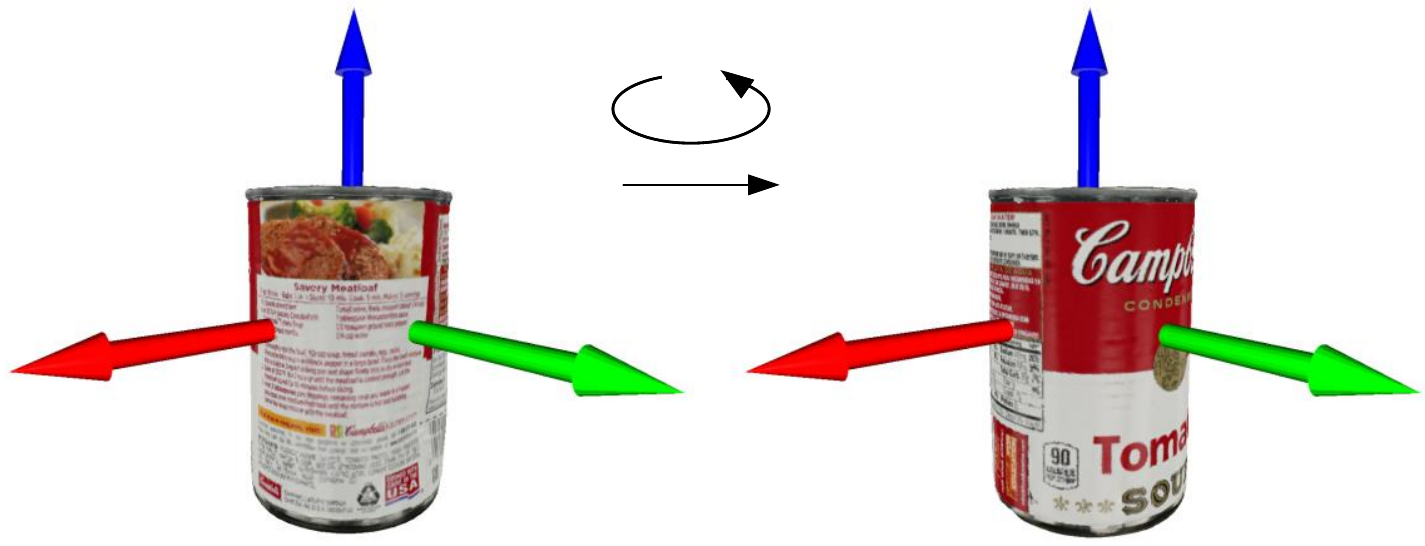}
    \caption{Counter-example of the violation of the first requirement. The object rotates for 180$^{\circ}$ around the z axis from the first pose to the second one.}
    \label{fig:violation1}
\end{figure}
As shown in Fig.~\ref{fig:violation1}, the object is a cylinder with asymmetric texture. The predicted pose rotates around the axis for $180^{\circ}$. According to the definition of ADD-S, 
\begin{equation}
    \label{eqn:violation_1}
    \mbox{ADD-S}(o, P_{gt}, P_{pred}) = 0,
\end{equation}
but $ P_{gt} \ne P_{pred}$.
\end{proof}
\begin{proof}
\emph{Violation of requirement 2.} 
\begin{figure}[htbp]
    \centering
    \includegraphics[width=0.4\linewidth]{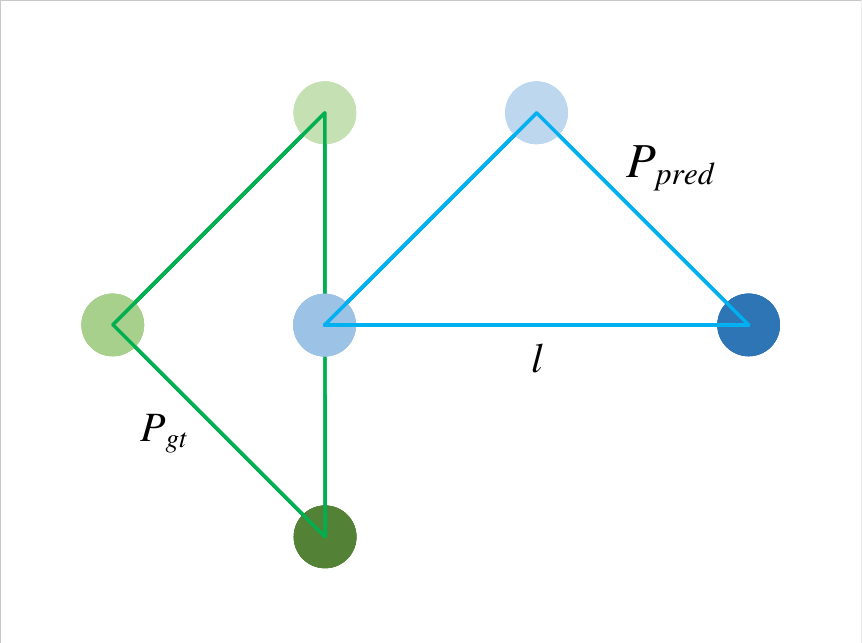}
    \caption{Counter-example of the violation of the second requirement.}
    \label{fig:violation2}
\end{figure}
As shown in Fig.~\ref{fig:violation2}, the object is only composed of three points forming an isosceles right triangle. The triangle has no pose ambiguity since each point has different color. The length of the longest edge of the triangle equals to $l$. The predicted pose and ground truth pose are perpendicular to each other. 
\begin{equation}
    \label{eqn:violation_2}
    \begin{split}
    \mbox{ADD-S}(o, P_{gt}, P_{pred}) &= \frac{l}{2} \\
    \mbox{ADD-S}(o, P_{pred}, P_{gt}) &= \frac{2 + \sqrt{5}}{6}l\\
    \end{split}
\end{equation}

As calculated in Equation~\ref{eqn:violation_2}. $\mbox{ADD-S}(o, P_{gt}, P_{pred}) \ne \mbox{ADD-S}(o, P_{pred}, P_{gt})$, so it doesn't meet the second requirement. Actually, in most of the situations, $\mbox{ADD-S}(o, P_{gt}, P_{pred}) \ne \mbox{ADD-S}(o, P_{pred}, P_{gt})$.
\end{proof}




\subsection{Metrics with Pose
Ambiguity}\label{subsec:metric_no_ambiguity}
In this situation, it is hard to give a formal statement on the rationality of metrics like Equation~\ref{eqn:r1},~\ref{eqn:r2} and~\ref{eqn:r3} as there are a set of ground truth poses.

However, the definition of ADD obviously doesn't make sense because it can only compare the predicted value to one ground truth pose.

For ADD-S, it is a compromise for the irrationality of ADD in this situation. It can evaluate the performance of 6D pose estimation to some extent. But in some cases, it also has some problems. One counter-example is given below.

\begin{figure}[htbp]
    \centering
    \includegraphics[width=0.6\linewidth]{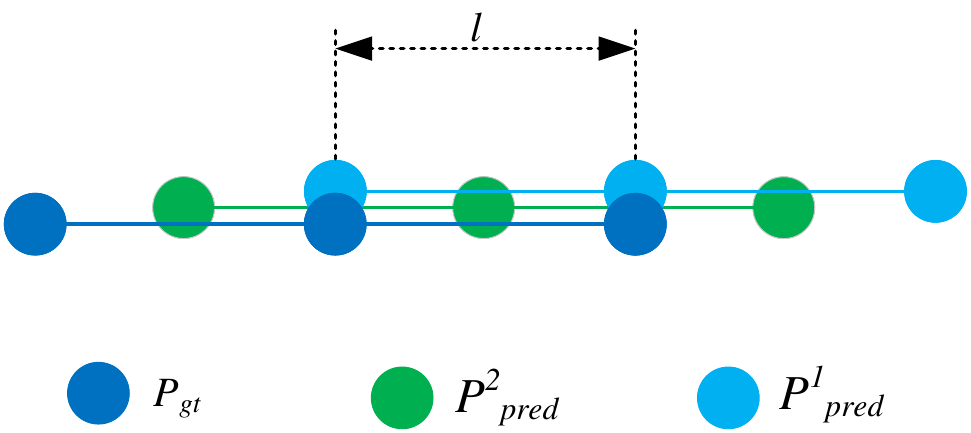}
    \caption{
    The three object actually align in the vertical direction. It is deliberately staggered in this direction for better visualization.}
    \label{fig:violation3}
\end{figure}

As shown in Fig.~\ref{fig:violation3}, $P_{pred}^1$ has greater pose estimation error than that of $P_{pred}^2$.
However, the ADD-S score of $P_{pred}^1$ is smaller than that of $P_{pred}^2$ as shown in Equation~\ref{eqn:result}.

\begin{equation}
    \label{eqn:result}
    \begin{split}
        \mbox{ADD-S}(o, P_{pred}^1, P_{gt}) = \frac{1}{3}l \\
        \mbox{ADD-S}(o, P_{pred}^2, P_{gt}) = \frac{1}{2}l \\ 
    \end{split}
\end{equation}
\begin{figure}[t]
    \centering
    \includegraphics[width=1.00\linewidth]{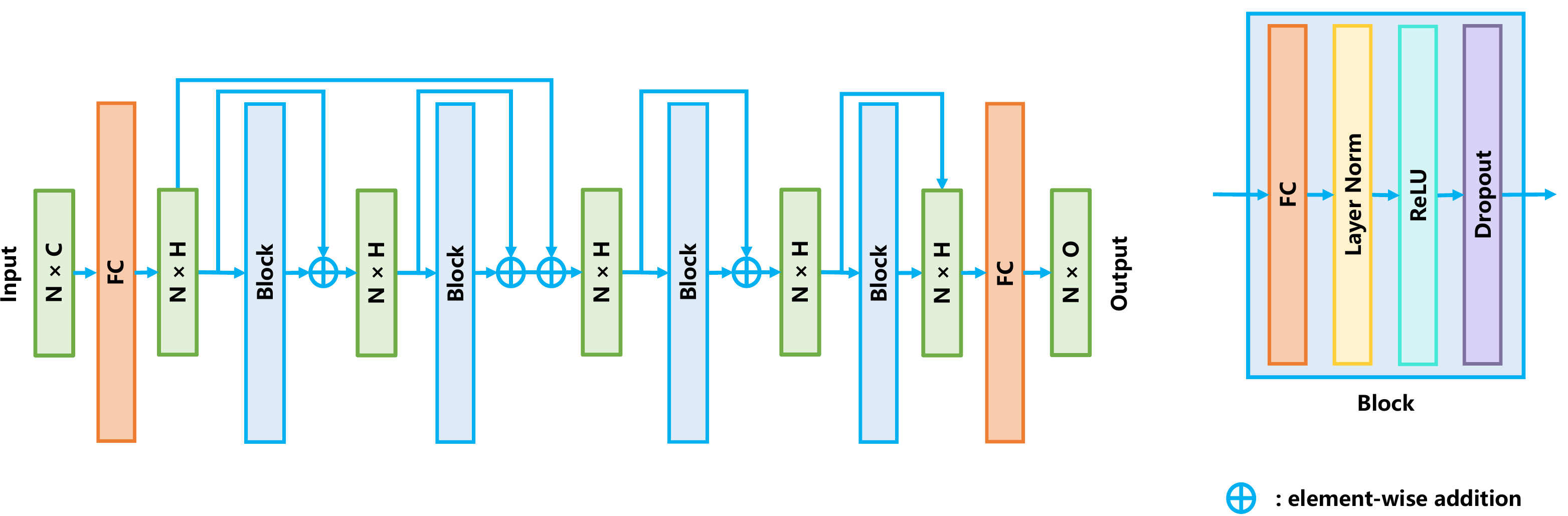}
    \caption{Structure of the Multi-Layer Perceptron (MLP) with residual blocks used in our networks. FC represents Fullly Connected Layer.}
    \label{fig:mlp_structure}
\end{figure} 
\subsection{Conclusion}
As discussed above, for the task of object 6D pose estimation, ADD is a suitable metric for objects without pose ambiguity but cannot be applied to those with pose ambiguity. ADD-S is problematic for objects without pose ambiguity and has some drawbacks for objects with pose ambiguity. As a result, neither ADD nor ADD-S can be used to evaluate objects pose for those both with pose ambiguity and without ambiguity in a unified manner. 
\subsection{MLP Structure}\label{sec:mlp_structure}

The stucture of the Multi Layer Perceptron (MLP) in the main paper is shown in Fig.~\ref{fig:mlp_structure}, which is composed of 4 blocks.

\end{document}